\pdfoutput=1

\documentclass[11pt]{article}

\usepackage[dvipsnames,svgnames]{xcolor}

\usepackage[preprint]{acl}

\usepackage{times}
\usepackage{latexsym}
\usepackage{booktabs}
\usepackage{marvosym}

\usepackage[T1]{fontenc}

\usepackage[utf8]{inputenc}

\usepackage{microtype}

\usepackage{inconsolata}

\usepackage{graphicx}

\usepackage{amsmath}
\usepackage{algorithm}
\usepackage{algorithmic}

\usepackage{subcaption} 
\usepackage{multirow} 
\usepackage{multicol}
\usepackage{colortbl}

\usepackage{enumerate}
\usepackage{enumitem}
\usepackage{amssymb}
\usepackage{listings}
\usepackage{minitoc}

\newcommand{\dyf}[1]{\textcolor{black}{#1}}

\definecolor{jwtcolor}{RGB}{0, 123, 255}
\newcommand{\jwt}[1]{{\color{jwtcolor}#1}}

\makeatletter
\def\thanks#1{\protected@xdef\@thanks{\@thanks\protect\footnotetext{#1}}}
\makeatother
\title{Dynamic Parallel Tree Search for Efficient LLM Reasoning}

\author{Yifu Ding\textsuperscript{1,2}, Wentao Jiang\textsuperscript{3}, Shunyu Liu\textsuperscript{2}, Yongcheng Jing\textsuperscript{2}, Jinyang Guo\textsuperscript{1}, Yingjie Wang\textsuperscript{2}, \\
{\bf Jing Zhang\textsuperscript{3}, Zengmao Wang\textsuperscript{3,*}, Ziwei Liu\textsuperscript{2}, Bo Du\textsuperscript{3}, Xianglong Liu\textsuperscript{1,*}, Dacheng Tao\textsuperscript{2,*}}
  \thanks{
  \textsuperscript{*}Corresponding author. \textsuperscript{1}Beihang University, China. 
  \textsuperscript{2}Nanyang Technological University, Singapore. 
  \textsuperscript{3}Wuhan University, China. }
  }

\begin{document}
\maketitle

\begin{abstract}
Tree of Thoughts (ToT) enhances Large Language Model (LLM) reasoning by structuring problem-solving as a spanning tree. However, recent methods focus on search accuracy while overlooking computational efficiency. The challenges of accelerating the ToT lie in the frequent switching of reasoning focus, and the redundant exploration of suboptimal solutions. To alleviate this dilemma, we propose Dynamic Parallel Tree Search (DPTS), a novel parallelism framework that aims to dynamically optimize the reasoning path in inference. 
It includes the Parallelism Streamline in the generation phase to build up a flexible and adaptive parallelism with arbitrary paths by fine-grained cache management and alignment. 
Meanwhile, the Search and Transition Mechanism filters potential candidates to dynamically maintain the reasoning focus on more possible solutions and have less redundancy. Experiments on Qwen-2.5 and Llama-3 with Math500 and GSM8K datasets show that DPTS significantly improves efficiency by 2-4$\times$ on average while maintaining or even surpassing existing reasoning algorithms in accuracy, making ToT-based reasoning more scalable and computationally efficient. 
\end{abstract}
\section{Introduction}

The advent of OpenAI-o1~\cite{jaech2024openai}, a reasoning large language model, has sparked significant interest in the academic community. A key factor behind its success is the Chain-of-Thought~(CoT)-based reasoning technique~\cite{Wei_2022_Chain, chu2023survey}, which improves model’s reasoning ability by breaking complex problems into explicit intermediate steps. Building upon this, the Tree of Thoughts~(ToT)~\cite{Yao_2023_Tree} framework has been introduced to further elevate LLMs’ reasoning capacities, and be widely used in multi-step reasoning tasks~\cite{plaat2024reasoning}. ToT restructures reasoning as a tree search process and employs search algorithms, such as Monte Carlo Tree Search~(MCTS)~\cite{chaslot2008monte,xie2024monte}, to construct a tree-like structure that explores various reasoning pathways, leading to more refined and accurate responses~\cite{sprueill2023monte}.
However, current ToT approaches predominantly focus on improving search accuracy while overlooking computational efficiency~\cite{xie2024monte, cheng2024self, zhao2024marco}. We conclude two significant challenges that complicate the acceleration of ToT. 

\begin{figure}
   \begin{subfigure}{\linewidth}
   \vspace{-0.1in}
    \includegraphics[width=\linewidth]{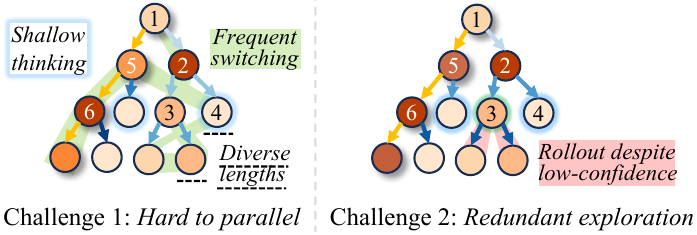}
    \caption{Challenges of implementing parallelism in reasoning tasks.}
    \label{fig:intro}
\end{subfigure}
\hfill
\begin{subfigure}{\linewidth}
    \includegraphics[width=\linewidth]{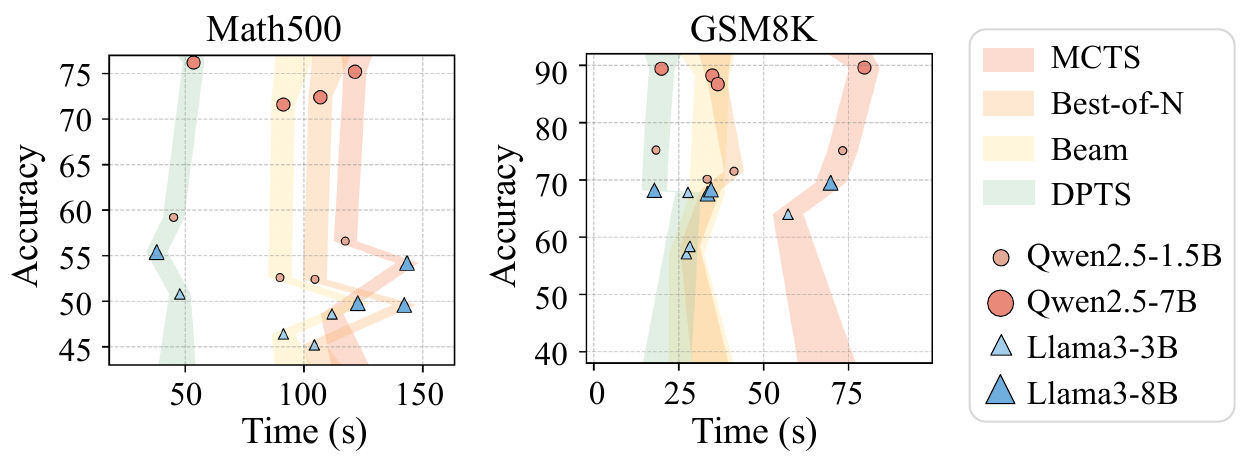}
    \caption{Accuracy and efficiency comparisons across tree search algorithms on various models and datasets.}
    \label{fig:comparison}
\end{subfigure}
\vspace{-0.2in}
\caption{(a) The demonstration of challenges. (b) The experimental results. }
\vspace{-0.25in}
\end{figure}



The first challenge arises from the frequent switching of reasoning focus in traditional sequential tree search, making it difficult to effectively parallelize~\cite{snell2024scaling}. 
Unlike conventional deep learning inference that is compatible with 
end-to-end parallelism, tree search has irregular computational trajectories. As shown in Figure~\ref{fig:intro} Challenge 1, it frequently switches among paths, including retrospect and recursion behaviors which complicates the parallelism (green trajectory). 
For example, if parallelizing nodes 6 and 2 in one generation, the different context lengths and KV cache require additional processing. 
Moreover, frequent switching 
also tends to yield shallow exploration~\cite{wang2025thoughts}. Due to the limited time and memory budgets, more explored paths mean less exploitation in deep paths. 



The second challenge stems from the redundant explorations on suboptimal solutions~\cite{li2024escape, besta2024topologies}. Previous tree search methods fail in identifying the less potential path and terminating it timely~\cite{wan2024dynamic}. Methods like MCTS attempt to balance the exploitation and exploration paths during node selection, but the selected nodes continue to roll out till the termination conditions (time or token limits) even with small prior confidence~\cite{xie2024monte}. For example, dark nodes in Figure~\ref{fig:intro} Challenge 2 have higher confidence. Node 3 with a lower confidence than node 5 will be explored earlier (pink trajectory). 
However, we observe that paths with low prior confidence have less probability of reaching the optimal solution, as illustrated in Section~\ref{sec:3.2}. 

To address these challenges, we propose a novel and efficient tree search framework, \textbf{DPTS (Dynamic Parallel Tree Search)}. 
This framework implements parallelized tree search and optimizes it by dynamically adjusting reasoning focus during the tree growing, thereby improving computational efficiency. 
It consists of two key components for both the generation and selection phases. 
(1)~DPTS implements a Parallelism Streamline tailored for LLM reasoning in the generation phase. It facilitates the rollout for arbitrary paths in parallel, allowing the expanded nodes to be rearranged at each step. 
Additionally, we carefully engineer cached data collection and context alignment, paving the way for parallelized inference with varying path length and node selection. 
(2)~Building on this, to prevent excessive exploitation and focus the reasoning on more potential paths,  
we introduce a Search and Transition Mechanism in the selection phase. 
It dynamically balances the exploitation-exploration paths by the bidirectional transition, i.e., \textit{Early Stop} (Exploitation → Exploration), \textit{Deep Seek} (Exploration → Exploitation), allowing the model to focus on the most promising solutions and mitigate inefficient exploitation on suboptimal solutions. 

Our work provides valuable insights into accelerating the ToT for LLM reasoning, paving the way for future work to solve real-world challenges. 
Our contributions can be concluded as follows:
\begin{itemize}[itemsep=0pt,parsep=2pt,topsep=3pt]
    \item We propose the DPTS framework, which solve the frequent switching and redundant exploration issues in previous tree search methods for LLM reasoning. 
    \item The Parallelism Streamline provides a flexible and efficient generation in node parallel, which bridges the gap between the sequential tree structure and parallelized inference. 
    \item The Search and Transition Mechanism exploits the most potential solutions and reduces unnecessary exploitation, ensuring that high-confidence nodes receive deeper reasoning. 
    \item Experiments shows that DPTS reaches the best solution with less inference time across various models and widely used reasoning datasets, as plotted in Figure~\ref{fig:comparison}. 
\end{itemize}

\section{Related Work}

\paragraph{Reasoning with LLMs. }
LLMs have evolved from System 1 tasks (e.g., translation)~\cite{Brown_2020_Language} to System 2 reasoning (e.g., math, logic)~\cite{Kojima_2022_Large}. CoT~\cite{Wei_2022_Chain} enhances multi-step reasoning, with variants like Self-Consistent CoT~\cite{wang2022self}, but its exploration scope remains constrained, limiting its effectiveness.~~\cite{chu2023survey}.
Furthermore, ToT~\cite{Yao_2023_Tree} enables multi-path exploration, leveraging MCTS~\cite{chaslot2008monte} for backtracking and heuristic rollouts~\cite{wan2024alphazero,wang2024q}. However, MCTS remains computationally expensive, with limited work on acceleration methods.

\paragraph{LLM Inference Acceleration. }
While LLM inference has been optimized for linear decoding~\cite{lin2024awq}, tree-structured reasoning remains underexplored~\cite{li2024large}. Approaches like Deft~\cite{yao2024deft} optimize prefix sharing, while others use self-consistency for early stopping~\cite{li2024escape}. Efficient tree search for LLM reasoning remains an open challenge.

Due to page limit, we have included a more detailed discussion of related work 
in Appendix~\ref{app:sec:related_work}. 
\section{Method Rationale}

\label{sec:motivation}

 

{
In this section, we present empirical findings that highlight the key challenges of tree search in LLM and provide the rationale behind our proposed DPTS. 
}
First, the frequent switch between paths complicates parallel execution and causes shallow thinking, disrupting the model’s ability to engage in efficient deep reasoning (Sec.~\ref{sec:3.1}). 
Second, excessive exploitation of low-confidence paths results in redundant rollouts and wastes effort on fewer possible candidates (Sec.~\ref{sec:3.2}). 

\subsection{Frequent Switching}
\label{sec:3.1}

\begin{figure}[t]
    \centering
    \includegraphics[width=0.93\linewidth]
    {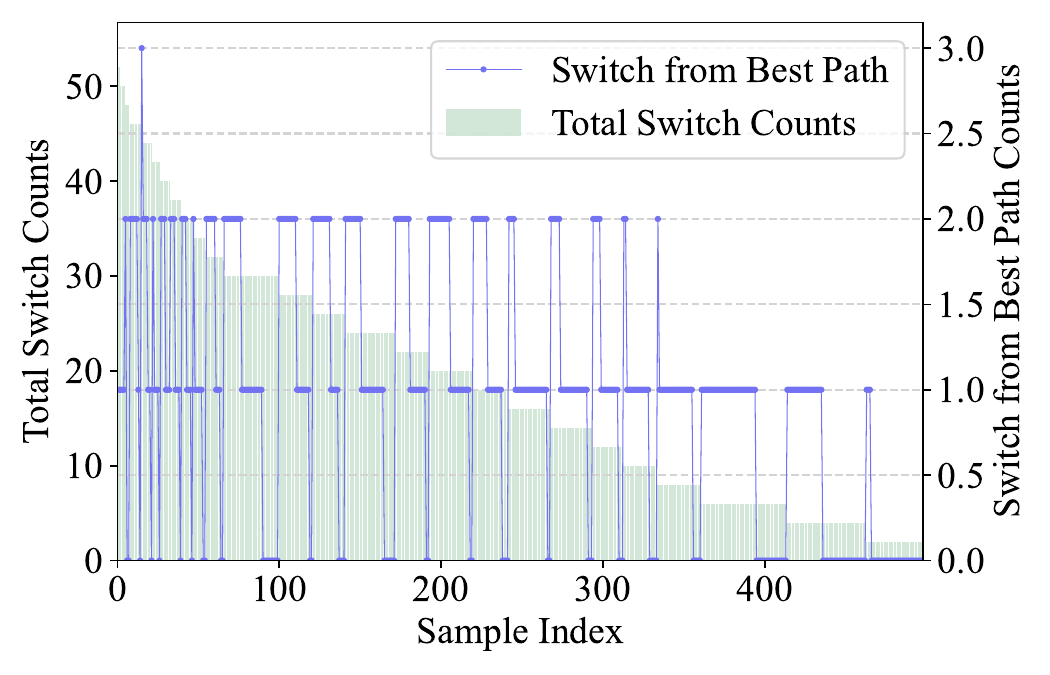}
    \vspace{-0.1in}
    \caption{Statistics for switch from the best path to the suboptimal (blue), and total switch (green).} 
    \vspace{-0.1in}
    \label{fig:motivation_switch_path}
\end{figure}


Tree search inherently exhibits retrospective and recursive behaviors, making efficient parallel execution difficult. Even if each node is constrained to generate the same number of tokens, the focus switching between different reasoning trajectories and the diverse path lengths makes it incompatible with the end-to-end parallelism on GPUs. The detailed illustrations for this phenomenon can be found in Appendix~\ref{sec:app:trees}. 


The focus switching between paths also makes the tree search fail in focused reasoning trajectory~\cite{wang2025thoughts}, which prevents deep thinking and leads to a tendency of shallow exploitation. We quantify the switch times of the reasoning focus on each sample in the Math500 dataset. Figure~\ref{fig:motivation_switch_path} counts the total switch, which is about 35 on average. As well as the switch from the best path to a suboptimal or incorrect one, which is up to 3 times for a single sample. It demonstrates the instability of the tree search algorithm in maintaining a focused reasoning trajectory.

\jwt{



}

\subsection{Redundant Exploration}
\label{sec:3.2}

\begin{figure}[t]
    \includegraphics[width=0.85\linewidth]{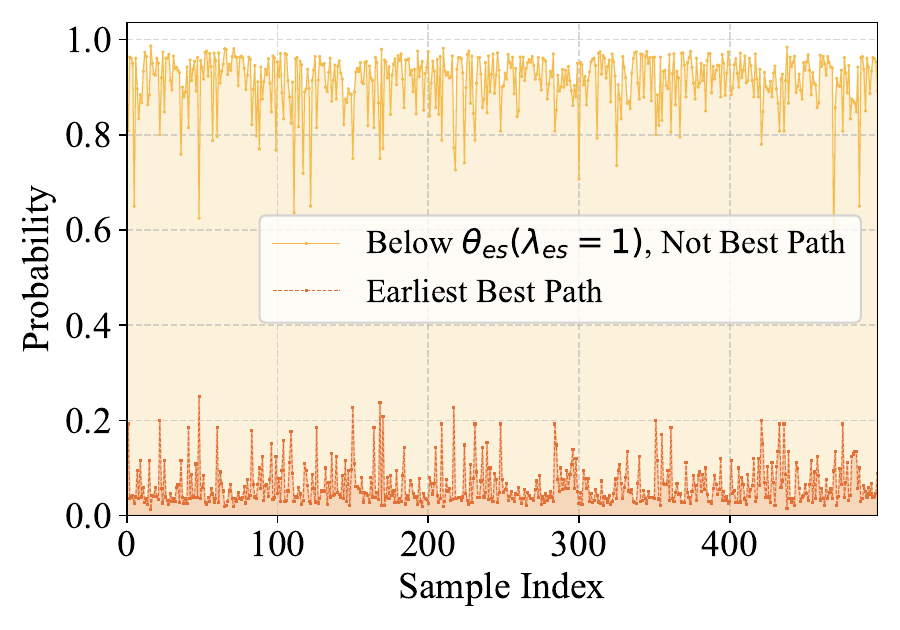}
    \vspace{-0.1in}
    \caption{Probabilities with reordered samples of those have prior confidence below $\theta_{es}(\lambda=1)$ in Eq.~\ref{eq:theta} and do not terminate with the highest reward score (yellow), and paths that are not the earliest best path (orange), which means there is already at least one path that has terminated with the same reward score. }
    \vspace{-0.2in}
    \label{fig:motivation_low_confidence}
\end{figure}

The lack of early termination in existing tree search algorithms leads to excessive exploitation and redundant searching. Observations in Figure~\ref{fig:motivation_low_confidence} show that low-confidence nodes rarely contribute to the best solutions, either terminated with suboptimal results (yellow) or failing to be the first to reach the best path (orange). The average probability of the suboptimal results brought by low confidence is 91.3\%, while the probability of those nodes being the earliest best path is only 6.2\%. It suggests that low-confidence nodes have little potential to reach the best solution, it is even hard to be the first one. It means that most low-confidence nodes have less contribution to the final results but waste computational resources. 
\jwt{


}

These findings emphasize the importance of maintaining the focus on deep reasoning and pruning low-confidence paths for efficient inference.

\section{Proposed Method}
\label{sec:method}
To address the aforementioned challenges, we propose an innovative framework that allows for efficient reasoning, termed Dynamic Parallel Tree Search (DPTS). 
In the generation phase, the Parallelism Streamline in Sec.~\ref{sec:method_parallel} supports fine-grained and flexible paralleled expansion for arbitrary paths. 
In the selection phase, the Search and Transition Mechanism in Sec.~\ref{sec:search_and_transition} enables less redundant exploration by identifying the highly potential solutions to focus reasoning.

\begin{figure*}
    \centering
    \includegraphics[width=\linewidth]{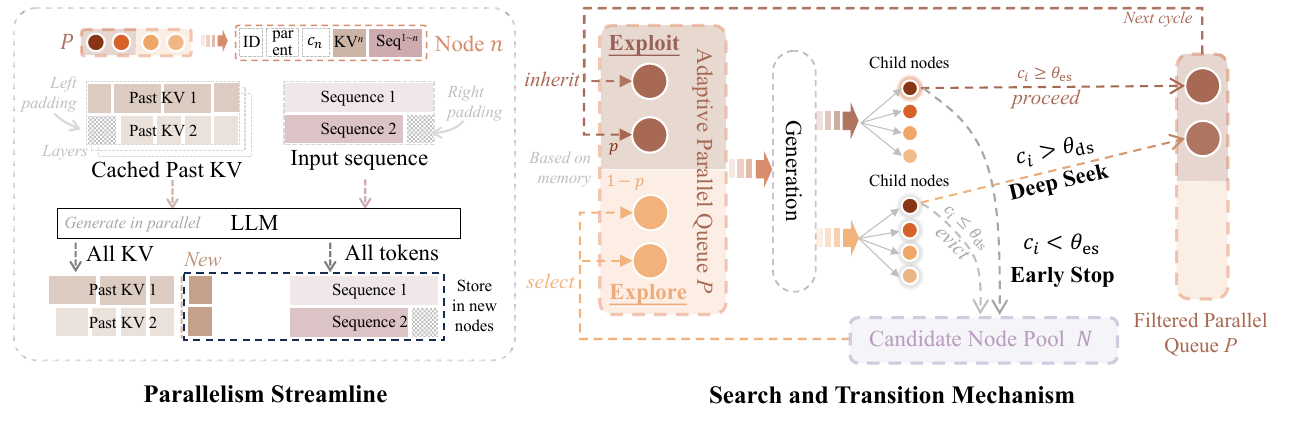}
    \vspace{-0.35in}
    \caption{Overview of the proposed DPTS framework. The right part demonstrates the Parallelism Streamline, while the left and middle illustrate the proposed Search and Transition Mechanism. }
    \label{fig:overview}
    \vspace{-0.1in}
\end{figure*}

\subsection{Parallelism Streamline}
\label{sec:method_parallel}

As illustrated in Figure~\ref{fig:overview}, We fully parallelize the tree search process in our framework with three main components: \textbf{Tree Structure Building}, \textbf{KV Cache Handling}, and \textbf{Adaptive Parallel Generation}. Each component is designed to optimize memory usage and parallel execution during the reasoning process.

\subsubsection{Tree Structure Building}
The tree search framework relies on a tree structure where each node represents a reasoning state.  Specifically, the node data structure includes the following elements:
\begin{itemize}[itemsep=0pt,parsep=0pt,left=0pt,topsep=0pt]
    \item \textbf{Node ID}: A unique identifier for each node.
    \item \textbf{Parent Node}: A reference to the parent node, establishing the hierarchical structure of the tree.
    \item \textbf{Prior Confidence}: The confidence of the node, based on prior knowledge and model predictions.
    \item \textbf{Key-Value Cache} (\(\text{KV}^n\)): The key-value cache specific to this node, storing intermediate results during the reasoning process.
    \item \textbf{Token Sequence} (\(\text{Seq}^{1 \sim n}\)): The complete token sequence from the root node to the current node, representing the reasoning path taken so far.
\end{itemize}
The key challenge lies in managing the KV cache~\cite{floridi2020gpt}. Instead of storing the entire sequences, each node only retains its own KV cache. This significantly reduces memory usage, particularly when dealing with a large number of nodes in the tree. By keeping each node's cache isolated, we avoid redundant memory usage while ensuring that each node has necessary information to continue reasoning process.

\subsubsection{KV Cache Handling}

The KV cache for each node is stored separately, and during parallel execution, these caches need to be collected and concatenated for efficient parallelism. The key challenge is that tree search paths have varying lengths, which means that both the KV caches and the input sequences for different nodes will vary in size and be hard to parallel. 

To address this, we use a simple but straightforward padding technique to ensure that all sequences have consistent lengths before being processed. Specifically, for nodes with shorter KV caches, we apply left padding with zeros. Similarly, input sequences are padded with a predefined padding token to match the longest sequence in the batch. 
This padding ensures that all nodes are processed in parallel with consistent sequence lengths and corresponding KV cache, allowing for efficient batch processing across the tree search. \dyf{The details of padding and concatenating are given in Appendix Eq. (\ref{eq:kv}) and (\ref{eq:seq}).}

Besides data collecting and preparation, we also clean up the useless KV cache either the leaf node is terminated, or all the children's branches are exploited and finished. In this way, we release the memory, making room for new reasoning paths. 

\subsubsection{Adaptive Parallel Generation}
To further utilize the computational resources, we introduce an adaptive parallelism queue, which dynamically adjusts the number of parallel paths based on the available GPU memory. The parallelism queue size, denoted \( |P| \), is used to restrict the number of exploitation and exploration paths in Sec.~\ref{sec:method_searching}. It is calculated by the available and the peak memory usage during previous generations: 
\begin{equation}
\label{eq:queue_size}
|P| = \frac{O_{\text{max}} - O_{\text{init}}}{O_{\text{peak}} - O_{\text{init}}},
\end{equation}
where $O_{\text{max}}$ is the total memory budget, \( O_{\text{peak}} \) represents the peak memory usage from the previous generation, and \( O_{\text{init}} \) is the memory consumption during model initialization. 
\dyf{As the tree grows, the memory occupation of intermediate results continues to increase even with KV cache cleaning. Since memory overflow is one of the termination conditions, it is important to adaptively adjust the parallel number, preventing excessive memory allocation and early termination.} 

After the generation phase, the newly generated sequences and KV caches are stored based on the tree width. The sequences for each node are completely stored, while the KV caches are stored partially with only the new tokens generated at this step (details are provided in Appendix~\ref{eq:n_new}). 
The new nodes are then added to the candidate node pool \( N \), where they will be available for subsequent selection processes in the tree search.

\vspace{0.05in}

In summary, our Parallelism Streamline is a well-structured streamline to optimize both memory usage and parallel execution. 
The overall process is showcased in Algorithm~\ref{alg:algorithm_parallel}. More details can be found in Appendix~\ref{app:sec:parallel_reasoning} due to the limited length.

\begin{algorithm}[ht]
\caption{Algorithmic process DPTS}
\label{alg:algorithm_parallel}
\begin{algorithmic}[1]
\REQUIRE LLM generation function $llm(x)$, PRM reward function $prm(x)$, Query $q$, Candidate Node Pool $N = \varnothing$, Parallel Queue $P = \varnothing$, Exploit Node Proportion $p$, Tree Width $w$. \\
\ENSURE End node with best path reward $n^*$.

{\color{ForestGreen}{// Step 1: Initialize the root node}} \\
\STATE $r \gets \text{generate\_node}(q, \mathrm{None})$
\STATE $N \gets N \cup \{r\}$

\WHILE{\text{within computational budget}} 

    \STATE $ P_\text{size} \gets \text{Eq. (\ref{eq:queue_size})}$  \\
    
    {\color{ForestGreen}{// Step 2: Perform searching}} \\
    \STATE $P \gets \text{Search}(P, P_\text{size}, N)$ (Algorithm~\ref{alg:searching})
    \STATE $\theta_{\mathrm{es}}, \theta_{\mathrm{ds}} \gets \text{Eq. (\ref{eq:theta})}$ \\
    
    {\color{ForestGreen}{// Step 3: Parallelism by Eq. (\ref{eq:kv}) and (\ref{eq:seq})}} \\
    \STATE $\mathbf{n} \gets \text{generate\_node}(
    \mathrm{Seq}^{all}, \mathrm{KV}^{all})$ \\ 
    {\color{ForestGreen}{// Step 4: Update new nodes by Eq.~(\ref{eq:n_new})}} \\
    {\color{ForestGreen}{// Step 5: Terminate and reward}} \\
    \STATE $N\gets \text{Reward}(P, N)$ (\text{Algorithm~\ref{app:algo:reward}}) \\
    {\color{ForestGreen}{// Step 6: Perform transition}} \\
    
    \STATE $P \gets \text{Transition}(P, \theta_{\mathrm{es}}, \theta_{\mathrm{ds}})$ (\text{Algorithm~\ref{alg:transition}})
\ENDWHILE

\RETURN $\max_{\text{reward}}(\forall n \in N)$
\end{algorithmic}
\end{algorithm}

\subsection{Search and Transition Mechanism}
\label{sec:search_and_transition}
In this section, we introduce the \textbf{Search} and \textbf{Transition} Mechanism in DPTS, which is a hybrid search algorithm that balances exploitation and exploration through separate management and dynamic conversion. 

\subsubsection{Search}
\label{sec:method_searching}




The Search Mechanism aims to balance exploration and exploitation by dynamically partitioning the nodes in parallel queue \( P \) into two categories: \textit{explore nodes} and \textit{exploit nodes}. 

As illustrated in Figure~\ref{fig:overview} (left), these nodes are selected from the candidate node pool \( N \).
The partition ratio $p$ can be manually adjusted according to the task and memory budget. 
At initialization, the top \( p|P| \) highest-scoring nodes are assigned as \textit{exploitation nodes}, while the remaining \( (1 - p)|P| \) nodes are assigned as \textit{exploration nodes}. While during searching progress, the proportion of the two types of nodes dynamically fluctuates based on the transition mechanism in Sec.~\ref{sec:method_transition}. 
The primary distinction between these two nodes lies in their origins and roles during the search process.

\paragraph{Exploitation Nodes} 
 The exploitation nodes are primarily inherited from parent exploitation nodes, focusing on refining the most promising paths in the search space. When a child node’s confidence exceeds a predefined threshold, it inherits the status of its parent exploitation node and continues that path. This inheritance ensures that the most promising paths are deepened and further refined. Additionally, when the number of exploitation nodes falls below a predefined threshold, new high-confidence candidate nodes from the pool \( N \) are selected to fill the gap, ensuring that the number of exploitation nodes remains adequate for the search process. This strategy enables the exploitation of high-potential paths while maintaining the focus on areas with high confidence.

\paragraph{Exploration Nodes} 
In contrast to exploitation nodes, the exploration nodes are not inherited from previous nodes but are dynamically selected from the candidate nodes pool. These nodes are responsible for discovering new paths that may have high potential but low current confidence in the search space. At each reasoning step, the exploration nodes are reselected from the candidate pool \( N \), choosing the highest-confidence nodes that are not already assigned as exploitation nodes. The dynamic re-selection of exploration nodes allows the search process to adapt to changing circumstances and uncover new regions of the search space that may lead to better solutions.




\subsubsection{Transition}
\label{sec:method_transition}

While the Search Mechanism ensures a balance between exploration and exploitation, the redundant issue is not entirely mitigated. 
One example is that the initial exploitation nodes are not guaranteed to be the optimal solution. However, they only stop exploiting till reach the termination condition. 
Another issue occurs when high-confidence nodes are assigned as exploration nodes, but they will wait for the computation resources and do not roll out till the previous paths terminate. 

To address these issues, we introduce the Transition Mechanism, which consists of two main strategies: \textit{Early Stop} (Exploitation → Exploration) and \textit{Deep Seek} (Exploration → Exploitation). As illustrated in Figure~\ref{fig:overview} (middle), these strategies allow an evolving search space with node transits between the two statuses. 
It helps the tree maintain focused reasoning, ensuring the efficient allocation and utilization of limited computational resources throughout the whole search process. 

\paragraph{Early Stop (Exploitation → Exploration)} 
The Early Stop~\cite{yao2007early} strategy allows relatively low-confidence exploitation nodes to transition into explore nodes, eliminating redundant exploitation on suboptimal paths. 
During the expansion process, if the best child node of an explore node has a confidence lower than a certain threshold \( \theta_{\mathrm{es}} \), the child node will be excluded from the queue \( P \) in the next cycle. This prevents further exploration of paths that are unlikely to lead to optimal solutions, saving computational resources. Conversely, if the child node’s confidence exceeds \( \theta_{\mathrm{es}} \), it inherits the parent’s status and continues to explore in the next cycle. This mechanism ensures that only the most promising explore nodes continue to expand, optimizing both exploration and resource usage.
The threshold \( \theta_{\mathrm{es}} \) is defined as follows:
\begin{equation}
\label{eq:theta}
\mathsf{\theta}_{\mathrm{es}} = \begin{cases}
\lambda_\mathrm{es} \, \frac{1}{|\mathcal{N}|} \sum\limits_{i \in \mathcal{N}} c_i, & \text{if } t \leq t^* \\
\underset{i \in \mathcal{N}}{\max} \, c_i, & \text{otherwise}
\end{cases}
\end{equation}
where \( \mathcal{N} \) is the set of previously expanded nodes, \( c_i \) represents the confidence of node \( i \), \( \lambda_\mathrm{es} \) is a coefficient that adjusts the threshold, \( t \) is the number of currently terminated paths, and \( t^* \) is a predefined threshold after which \( \theta_{\mathrm{es}} \) is adjusted.

\paragraph{Deep Seek (Exploration → Exploitation)} 
The Deep Seek strategy addresses the issue of inefficient over-exploration and shallow thinking, ensuring promising exploration nodes can be dug deeper. Specifically, exploration nodes with confidence exceeding a threshold \( \theta_{\mathrm{ds}} \) with $\lambda_\mathrm{ds}$ are promoted to exploitation nodes. 
As a result, the number of exploitation nodes may temporarily exceed \( p|P| \). But as more high-confidence nodes are promoted, \( \theta_{\mathrm{es}} \) increases, and thus more exploration nodes are stopped under the Early Stop strategy. This creates a dynamic balance between exploration and exploitation throughout the search process. 

\vspace{0.07in}

In a word, the proposed Search and Transition Mechanism in DPTS effectively manages the trade-off between exploitation and exploration with dynamic and bidirectional transition. 
Detailed algorithms in this part can be found in Appendix~\ref{app:sec:parallel_reasoning}.  


\section{Experiments}
\label{sec:exp}
We conduct extensive experiments to evaluate the efficiency of the DPTS framework. We benchmark its performance against various search algorithms across multiple models and datasets to ensure a comprehensive analysis.

\subsection{Settings}
\label{sec:exp_setting}

\paragraph{Models.} We include Qwen-2.5-1.5B-Instruct, Qwen-2.5-7B-Instruct~\cite{yang2024qwen2}, LLaMA-3.1-8B-Instruct, and LLaMA-3.2-3B-Instruct~\cite{touvron2023llama} to cover various model sizes.  

\paragraph{Datasets.} The evaluation datasets include Math500~\cite{hendrycks2021measuring} and GSM8K~\cite{cobbe2021training}, both are widely used for reasoning and mathematical problem-solving tasks. We implement the evaluation by referring the approaches used in Qwen-2.5-Math~\cite{yang2024qwen2}, supplemented in our submission. 

\paragraph{Comparison Methods.} We compare DPTS against three widely used search algorithms: Monte Carlo Tree Search~(MCTS)~\cite{sprueill2023monte}
, Best-of-N~\cite{cobbe2021training}
, Beam Search~\cite{Yao_2023_Tree}.
Since efficient tree search algorithms have recently regained attention after the emergence of LLM reasoning, the strong baselines are limited. As a result, we primarily compare DPTS against these typical and well-established search algorithms to demonstrate the effectiveness of our method. An introduction of the comparison methods and other details about experimental settings are provided in Appendix~\ref{app:sec:exp}.


\subsection{Comparisons on Search Algorithms}

\begin{table}[ht]
    \centering
    \footnotesize
    \caption{Comparisons across existing search algorithms on LLM reasoning tasks.}
    \vspace{-0.1in}
    \label{tab:comparisons}
    \begin{tabular}{@{}c@{\hskip 4pt} l@{\hskip 6pt} c@{\hskip 6pt} r c@{\hskip 6pt} r}
        \toprule
        \multirow{2}{*}{\textbf{Model}} & \multirow{2}{*}{\textbf{Algo.}} & \multicolumn{2}{c}{\textbf{Math500}} & \multicolumn{2}{c}{\textbf{GSM8K}} \\
        & & \textbf{Acc.} & \textbf{Time (s)} & \textbf{Acc.} & \textbf{Time (s)} \\
        \midrule
        \multirow{4}{*}{\begin{tabular}{c} Qwen-2.5 \\ 1.5B \end{tabular}} & MCTS & 56.6 & 117.37 & 75.1 & 73.28 \\
        & Best-of-N & 52.6 & 89.87 & 70.1 & 33.37 \\
        & Beam & 52.4 & 104.58 & 71.5 & 41.27 \\
        & \cellcolor[HTML]{D3D3D3} \textbf{DPTS} & \cellcolor[HTML]{D3D3D3} 59.2 & \cellcolor[HTML]{D3D3D3} \textbf{45.10} & \cellcolor[HTML]{D3D3D3} 75.2 & \cellcolor[HTML]{D3D3D3} \textbf{18.32} \\
        \midrule
        \multirow{4}{*}{\begin{tabular}{c} Qwen-2.5 \\ 7B \end{tabular}} & MCTS & 75.2 & 121.46 & 89.6 & 79.68 \\
        & Best-of-N & 71.6 & 91.29 & 88.2 & 34.89 \\
        & Beam & 72.4 & 106.89 & 86.7 & 36.49 \\
        & \cellcolor[HTML]{D3D3D3} \textbf{DPTS} & \cellcolor[HTML]{D3D3D3} 76.2 & \cellcolor[HTML]{D3D3D3} \textbf{53.50} & \cellcolor[HTML]{D3D3D3} 89.4 & \cellcolor[HTML]{D3D3D3} \textbf{19.95} \\
        \midrule
        \multirow{4}{*}{\begin{tabular}{c} Llama-3 \\ 3B \end{tabular}} & MCTS & 48.6 & 111.80 & 64.0 & 57.19 \\
        & Best-of-N & 46.4 & 91.34 & 57.1 & 27.27 \\
        & Beam & 45.2 & 104.36 & 58.4 & 28.27 \\
        & \cellcolor[HTML]{D3D3D3} \textbf{DPTS} & \cellcolor[HTML]{D3D3D3} 50.8 & \cellcolor[HTML]{D3D3D3} \textbf{47.75} & \cellcolor[HTML]{D3D3D3} 67.8 & \cellcolor[HTML]{D3D3D3} \textbf{27.74} \\
        \midrule
        \multirow{4}{*}{\begin{tabular}{c} Llama-3 \\ 8B \end{tabular}} & MCTS & 54.2 & 143.36 & 69.5 & 69.74 \\
        & Best-of-N & 49.8 & 122.63 & 67.6 & 33.48 \\
        & Beam & 49.6 & 142.21 & 68.3 & 34.51 \\
        & \cellcolor[HTML]{D3D3D3} \textbf{DPTS} & \cellcolor[HTML]{D3D3D3} 55.4 & \cellcolor[HTML]{D3D3D3} \textbf{37.98} & \cellcolor[HTML]{D3D3D3} 68.2 & \cellcolor[HTML]{D3D3D3} \textbf{17.82} \\
        \bottomrule
    \end{tabular}
    \vspace{-0.2in}
\end{table}


We conduct a comprehensive comparison across different search algorithms on various models and sizes. We emphasize the search efficiency of our method while maintaining accuracy. 

For efficiency, results in Table~\ref{tab:comparisons} show that DPTS significantly reduces inference time compared to other search methods across various models and datasets, demonstrating superior efficiency. 
On Math500, DPTS achieves the lowest inference time across all models. Particularly, in Qwen-2.5, DPTS reduces the search time from 117.37s (MCTS) to 45.10s in the 1.5B model, achieving nearly a $2.6\times$ speedup, and reduces from 121.46s (MCTS) to 53.50s in 7B model, accelerating nearly $2.2\times$. 
The impact is even more pronounced on the GSM8K, where DPTS achieves a $3.9\times$ speedup from 79.68s to 19.95s in Qwen-2.5-7B. And DPTS even only requires 17.82s for each sample using Llama-3-8B, also $3.9\times$. It forcefully suggests that DPTS effectively mitigates redundant rollouts and optimizes search efficiency. 
We highlight that, especially on the more challenging tasks, the Early Stop plays a crucial role. Without it, trees often run till timeout on Math500, significantly increasing inference time. In contrast, our approach allows the search tree to terminate earlier within a limited number of expansions, effectively reducing computation time. 

For accuracy, DPTS maintains the searching quality and even outperforms the existing algorithms with half or even less of the reasoning time. 
On the Math500 dataset, DPTS achieves the highest accuracy in all experiment cases, surpassing MCTS, Best-of-N, and Beam search. Notably, for Qwen-2.5-1.5B, DPTS improves accuracy from 56.6\% (MCTS) to 59.2\%. 
A similar trend is observed on GSM8K, where DPTS either matches or slightly improves accuracy over MCTS, and surpasses Best-of-N and Beam Search by a wide margin. On Llama-3-3B, DPTS has 67.8\% accuracy, outperforming the previous best MCTS by 3.8\% with only 48.5\% time consumption. 
These results highlight that DPTS maintains or even enhances solution quality while significantly improving inference speed, making it a more robust and efficient search algorithm for complex reasoning tasks.

\begin{table}
    \centering
    \caption{Ablation study of each component in DPTS framework. ``AP'': adaptive parallelism. ``S'': Searching. ``T'': Transition. ``Best Index'': The average index of terminated path leads to the best solution.  }
    \vspace{-0.1in}
    \label{tab:ablation}
    \resizebox{0.95\linewidth}{!}{
    \begin{tabular}{lccrc}
    \toprule
        \textbf{Algo.} & \textbf{${|P|}$} & \textbf{Acc.}  & \textbf{Time (s)} & \textbf{Best Index} \\ \midrule
         Baseline & 1 & 56.6 & 117.37 & 10.45 \\ \midrule
         Baseline & AP & 58.8 & 108.06 & 8.27 \\ 
         + S & AP & 58.2 & 76.81 & 4.66 \\ 
         + T & AP & 57.0 & 32.22 & 2.51 \\ 
         + S + T & AP & 59.2 & 45.10 & 4.17 \\ 
     \bottomrule
    \end{tabular}
    }
    \vspace{-0.2in}
\end{table}

\subsection{Ablation Study}

To analyze the contribution of each component within the DPTS framework, we conduct an ablation study on Qwen-2.5-1.5B with Math500 in Table~\ref{tab:ablation}. In this study, we use the classical MCTS as the baseline and incrementally integrate our proposed techniques to evaluate their impact. 

We begin with the original MCTS (non-parallel, $|P|=1$) as the baseline. It spends the most time per sample and has the largest best index 10.45. 
We then apply Parallelism Streamline with Adaptive Parallel Generation (AP), and accuracy improves. It shows that the trees are able to grow faster and larger to include a better solution with parallelism. 

Next, we assign the node status as exploit or explore nodes for each expansion with Search Mechanism, denoted as ``+S | AP'' in Table~\ref{tab:ablation}. 
The search process becomes significantly more structured and targeted, leading to a boost in efficiency. The time of each sample saves by 31.24s (28.9\%$\downarrow$). It finds the best path within an average of 4.66 terminated paths, much fewer than Baseline AP. 

Moreover, when only applying the Early Stop strategy in the Transition Mechanism (denoted as ``+T | AP), we obtain fast inference with much less time and paths. However, since we only use the exploitation nodes without exploring the possible branches, the accuracy is relatively low. Therefore, we claim that the Search and Transition Mechanism should be used as a whole: the Search mechanism provides different node statuses for exploitation and exploration, while the Transition mechanism makes them flexibly change and update. 

Finally, we combine our Search and Transition Mechanism (denoted as ``+S+T | AP''), enabling Early Stop and Deep Seek. It shows the best search results in accuracy and efficiency. 
It demonstrates that DPTS is efficient in quickly identifying optimal solutions and conducting deep reasoning. 

Results show that each component of DPTS contributes significantly to improving inference speed and reasoning accuracy, making it a robust and scalable framework for parallel tree search.

\subsection{Hyperparameter Analysis}

\begin{table}
    \centering
    \caption{Hyperparameter $\lambda_\mathrm{es}$ and $\lambda_\mathrm{ds}$ in transition thresholds $\theta_\mathrm{es}$ and $\theta_\mathrm{ds}$. ``ES (Early Stop)~\%'' and ``DS (Deep Seek)~\%'' are the ratios of the node type transition between the exploitation and exploration. More results can be found in Appendix~\ref{sec:app:hyperparameter}. }
    \vspace{-0.05in}
    \label{tab:hyperparameter}
    \resizebox{\linewidth}{!}{
    \begin{tabular}{cccccc}
    \toprule
        \textbf{$\lambda_\mathrm{es}$} & \textbf{$\lambda_\mathrm{ds}$} & \textbf{Acc.} & \textbf{Time (s)} & \textbf{ES~(\%)}  & \textbf{DS~(\%)} \\ \midrule
        1.0 & 1.0 & 53.0 & 47.59 & 41.4 & 10.5 \\ 
        0.9 & 0.9  & 58.6 & 43.30 & 15.6 & 20.9 \\
        0.8 & 0.8 & 58.0 & 46.33 & 8.1  & 23.9 \\
        0.6 & 0.6 & 57.4 & 44.39 & 6.1 & 32.3   \\
        0.4 & 0.4 & 56.6 & 38.41 & 0 & 0.1 \\
        \bottomrule
    \end{tabular}
    }
    \vspace{-0.12in}
\end{table}

We conduct a hyperparameter study in Table \ref{tab:hyperparameter} on the thresholds $\theta_\mathrm{es}$ and $\theta_\mathrm{ds}$ in the Transition mechanism. 
When $t < t^*$, the threshold $\theta$ follows the mean-based strategy determined by $\lambda$. When $t \geq t^*$, it turns to a max-based one. 
Empirically, we set  $t^* = 5$  to balance the flexibility and efficiency. 

Experimental results demonstrate that our method is robust to $\lambda$. We try different $\lambda$ and report the average ES/DS ratios per sample. 
We highlight that $\lambda_\mathrm{es}$ and $\lambda_\mathrm{ds}$ can be set differently based on the specific task. But DPTS consistently works well when $\lambda\in[0.6, 0.8]$. It demonstrates that the Transition mechanism is effective in mitigating the redundancy issue during search progress. 
However, it should be noticed that, if $\lambda$ is large, the ES transition may be aggressive, which leads to unsatisfactory results (e.g. $\lambda=1.0$). Meanwhile, if $\lambda$ is too small, it degrades to all exploitation nodes, resulting in low efficiency as well. 

\begin{figure}
    \centering
    \includegraphics[width=0.95\linewidth]{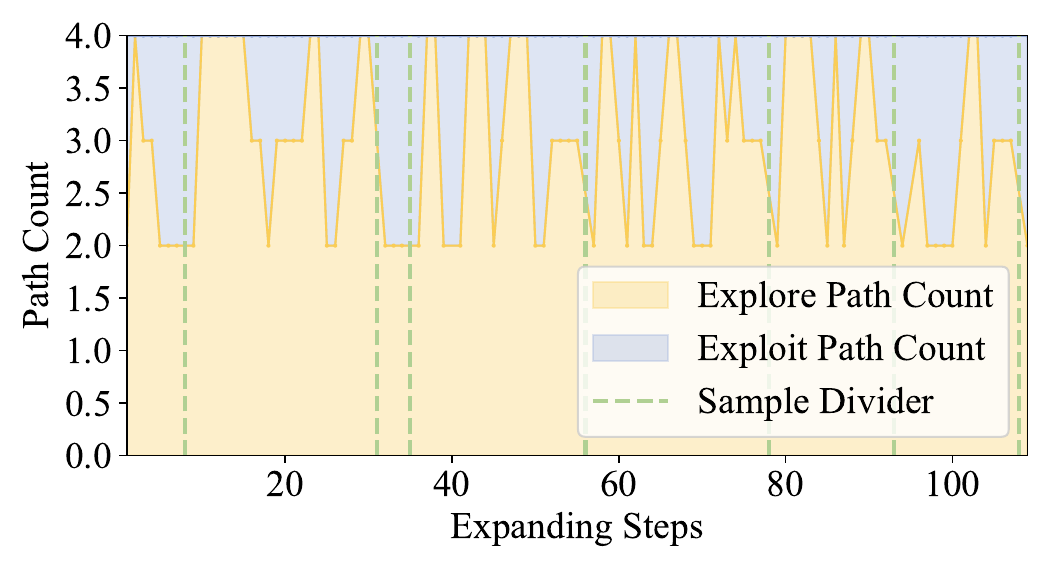}
    \vspace{-0.1in}
    \caption{Proportions of exploit and explore nodes throughout the search process. }
    \label{fig:ee_proportion}
    \vspace{-0.12in}
\end{figure}

\subsection{Visualizations}

To provide an intuitive understanding of the effectiveness of our proposed method, we present visualizations of searching trajectories. 

First, we analyze the dynamic changes in the number of exploitation and exploration nodes throughout the search process in Figure~\ref{fig:ee_proportion}. The Deep Seek transition temporarily increases the proportion of exploit paths, allowing promising nodes to receive deeper reasoning. However, as the threshold $\theta_\mathrm{es}$ increases, exploit nodes are more likely to reach the threshold and stop. As a result, the number of exploit nodes naturally decreases, reinforcing a balance between exploitation and exploration. This dynamic adaptation ensures that DPTS stretches on the most promising branches under the constraint of computational resources. 

\begin{figure}
    \centering
    \includegraphics[width=\linewidth]{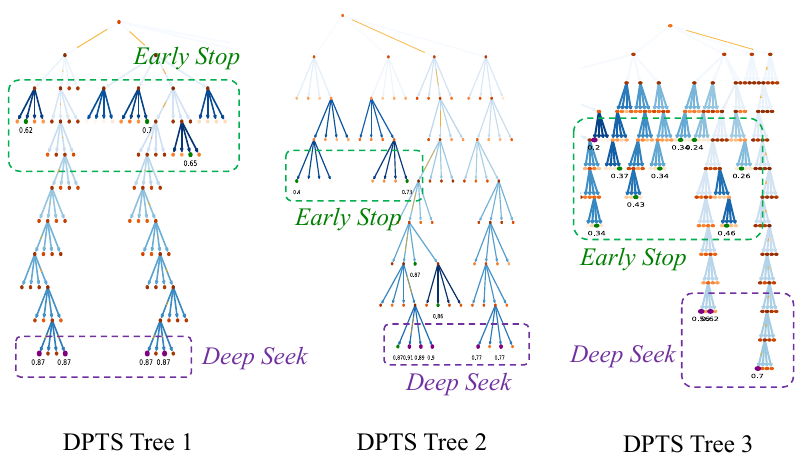}
    \vspace{-0.2in}
    \caption{Visualization of DPTS Tree. The green boxes are early stopped nodes based on their prior confidence using our \textit{Early Stop} mechanism, and the purple boxes are the terminated nodes with posterior reward scores. }
    \vspace{-0.12in}
    \label{fig:dpts_tree}
\end{figure}

Second, we show the trees generated by DPTS and analyze the search behavior in Figure~\ref{fig:dpts_tree}. It does not continue exploitation on low-confidence nodes, effectively pruning unpromising branches after shallow exploration. 
Additionally, the trees are capable of stable reasoning focus, with deep exploitation on promising paths. 
Therefore, the generated trees exhibit a relatively narrow width, as DPTS primarily expands nodes that are more relevant to the optimal path and spend less time on unnecessary regions. 
It demonstrates that DPTS 
ensures high-potential paths receive deeper thinking within a limited time and memory budget.

\section{Conclusion}


In this paper, we propose DPTS (Dynamic Parallel Tree Search) that effectively enhances the computational efficiency of LLM reasoning. DPTS introduces Parallelism Streamline, allowing efficient inference with arbitrary path lengths and nodes. The Dynamic Search and Transition Mechanism mitigates redundant rollouts and focuses on promising solutions. Experiments show that DPTS achieves the highest reasoning accuracy while delivering 2-4$\times$ speedup. 
Our work offers valuable insights into optimizing computation for efficient LLM reasoning, strengthening problem-solving capabilities to tackle real-world challenges.

\newpage 
\section*{Limitations}






Our DPTS framework focuses on selecting and refining the search paths, but does not involve hardware design. Therefore, it is orthogonal to low-level methods. For example, we can integrate DEFT~\cite{yao2024deft} to reduce the data transportation of the shared prefixes, leading to further acceleration. 
Also, our method is validated only on math reasoning tasks, and has not been tested on other domains, such as coding or scientific problems. However, we believe its generalizable capabilities make it applicable across a wide range of fields. 
Additionally, we envision that this method can also be applied to online training by improving generation quality. We leave these attempts to our future work.



\bibliography{acl_latex}

\appendix
\newpage

\section*{Appendix}
\label{sec:appendix}

\subsection*{Contents}
\begin{description}
    \item [\textbf{A}] \textbf{More Observations of Motivation} .............  \pageref{sec:app:motivation}
        \begin{description}
            \item [A.1] Wasted Tokens and Expansions ......... \pageref{sec:app:wasted}
            \item [A.2] Examples of DFS and BFS Trees ...... \pageref{sec:app:trees}
        \end{description}
    \item [\textbf{B}] \textbf{Formulas and Algorithms} ........................... \pageref{sec:app:formulas_and_algorithms}
        \begin{description}
            \item[B.1] Formulas in Parallelism Streamline ... \pageref{app:sec:parallel_reasoning}
            \item[B.2] Algorithms for Searching and Transition Mechanism ....................................... \pageref{sec:app:algorithms}
        \end{description}
    \item [\textbf{C}] \textbf{Additional Details about Experiment} ...... \pageref{app:sec:exp}
        \begin{description}
            \item [C.1] Comparison Methods ......................... \pageref{app:sec:comparison_methods}
            \item [C.2] Experimental Settings ....................... \pageref{app:sec:exp_setting}
            \item [C.3] Distribution of Best Path Index ......... \pageref{app:sec:best_path_index}
            \item [C.4] Additional Results of $\lambda$ ...................... \pageref{sec:app:hyperparameter}
        \end{description}
    \item [\textbf{D}] \textbf{Related Work} ............................................. \pageref{app:sec:related_work}
\end{description}

\hspace{0pt}

\section{More Observations of Motivation}
\label{sec:app:motivation}
\subsection{Wasted Tokens and Expansions}
\label{sec:app:wasted}

\begin{figure}[ht]
    \centering
    \includegraphics[width=\linewidth]{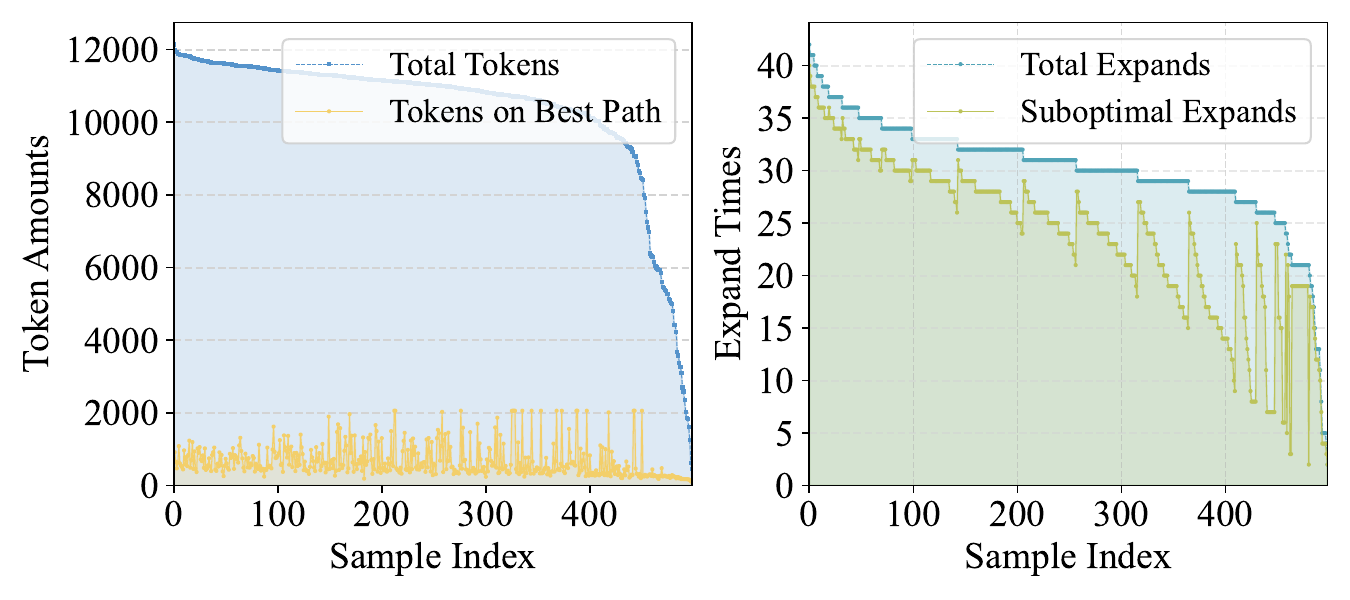}
    \caption{The proportion of tokens required for the best path relative to the total tokens generated (left), and the proportion of expansions on suboptimal paths relative to the total number of expansions (right). }
    \label{fig:motivation_waste_tokens}
\end{figure}

To better understand the inefficiencies caused by frequent node switching, we conduct a statistical analysis on Qwen-2.5-1.5B with the Math500 dataset and evaluate the redundancy in token generation and node expansion. 

Token redundancy analysis: In Figure~\ref{fig:motivation_waste_tokens}(left), we compare the total number of tokens generated for each sample (blue line) against the number of tokens required for the best path (yellow line). The samples are sorted in descending order primarily by total token count and secondarily by best-path token count. Our analysis shows that the total token count does not exhibit a strict multiplicative relationship with the best-path token count, but in general, the number of tokens required for the best path is significantly lower than the total token count.
This suggests that traditional tree search algorithms generate a large number of unnecessary tokens during exploration.

Expansion redundancy analysis: We also examined the number of node expansions during tree growth (Figure~\ref{fig:motivation_waste_tokens}(right)). The blue line represents the total number of expansions for each sample, while the green line represents the number of expansions on suboptimal paths (i.e., nodes that do not contain any part of the optimal solution). While there is no strict multiplicative correlation between these two metrics, the green line closely follows the blue line, indicating that a significant proportion of expansions occur on suboptimal paths. This further supports the observation that traditional tree search algorithms frequently explore unnecessary areas before finding the best solution.

\subsection{Examples of DFS and BFS Trees}
\label{sec:app:trees}

\begin{figure}[ht]
    \centering
    \includegraphics[width=\linewidth]{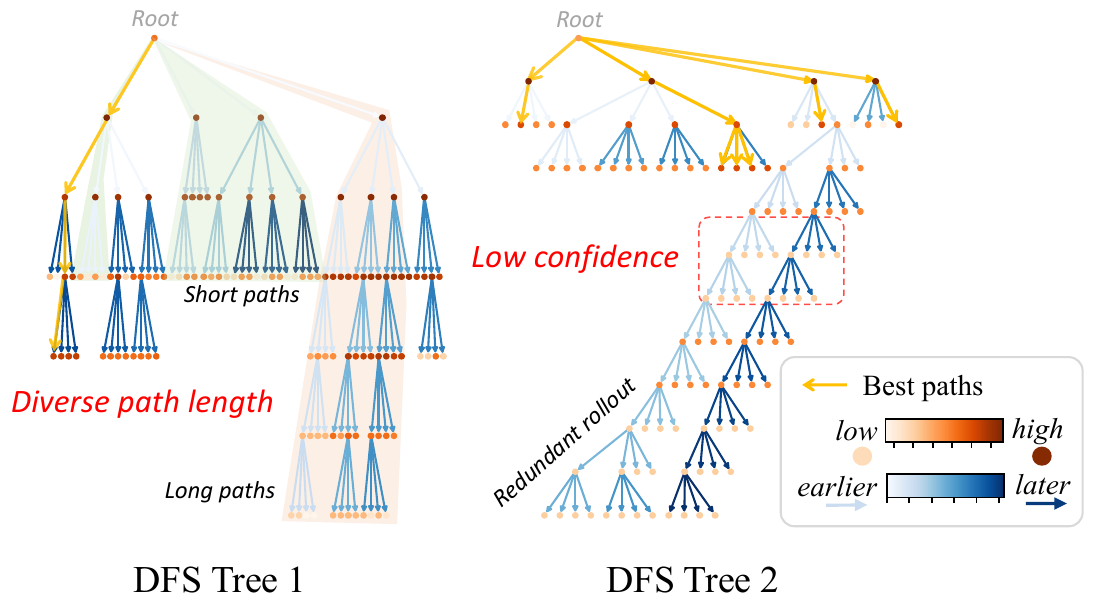}
    \caption{Growth of two trees with DFS algorithm. }
    \label{fig:motivation_dfs_trees}
\end{figure}

As illustrated in Figure~\ref{fig:motivation_dfs_trees}, which shows two typical depth-first search (DFS) trees, we visualize the node expansion process in layers based on tree depth. 
The darker reddish-brown nodes represent high-confidence nodes, while the lighter nodes indicate lower-confidence ones. The arrows denote parent-child relationships, where dark blue arrows indicate later-generated nodes and light blue arrows represent earlier-generated nodes. 

From the figure, we can clearly observe the reasoning trajectory of tree search: starting from the root node, the search prioritizes the child node with the highest confidence, then recursively expands deeper by selecting the most promising child node at each level. This continues until a termination condition is met, at which point the search backtracks and explores alternative paths from the root node. Due to the nature of this process, different paths vary significantly in their depth and termination points. Moreover, the next explored path does not follow a strict spatial or hierarchical pattern within the tree.

We also observe redundant exploration issues in the right two branches. At tree depths 4/5, the confidence scores of the expanded nodes are noticeably lower compared to previously explored nodes. However, due to the inherent mechanics of depth-first search (DFS), the algorithm continues expanding these nodes until the termination condition is met, even if the intermediate confidence scores remain consistently low. As a result, considerable computation is wasted on redundant expansions and token generations, with little contribution to improving the final output quality.

\begin{figure}[ht]
    \centering
    \includegraphics[width=\linewidth]{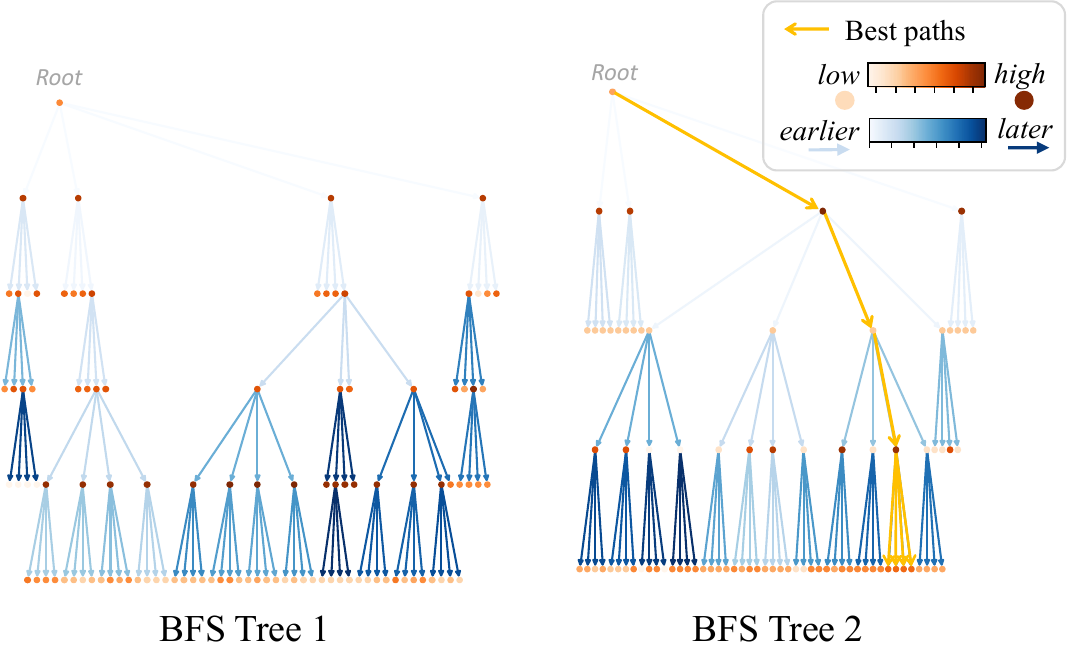}
    \caption{Growth of two trees with BFS algorithm. }
    \label{fig:motivation_bfs_trees}
\end{figure}


As illustrated in Figure~\ref{fig:motivation_bfs_trees}, BFS results in a flatter, more uniform, top-down expansion structure compared to the trees observed in Figure~\ref{fig:motivation_dfs_trees}. This behavior creates two key inefficiencies: 
(1) Incomplete reasoning before termination: In our experiments on the Math500 dataset, a pure BFS approach resulted in 178 (about 35.6\%)  of reasoning paths terminating without generating an answer (e.g., Tree 1 in Figure~\ref{fig:motivation_bfs_trees}). The algorithm explores many different areas of the tree but often fails to pursue any one path deeply enough to reach a valid conclusion. 
(2) Excessive expansions and token redundancy: Even when BFS eventually finds a correct answer, it tends to consume significantly more expansions and tokens than necessary (e.g., Tree 2 in Figure~\ref{fig:motivation_bfs_trees}). The best path (highlighted in yellow arrows) has a depth of only 4, yet before discovering this optimal solution, BFS explores a large number of additional nodes (light blue arrows), many of which do not contain any part of the optimal path.

\section{Formulas and Algorithms}
\label{sec:app:formulas_and_algorithms}

\subsection{Formulas in Parallelism Streamline}
\label{app:sec:parallel_reasoning}



\paragraph{Data Collection and Preparation.} 
Before executing parallel inference, we should collect the input data that is stored in separate memory locations. Based on the Node Data Structure (refer to Appendix~\ref{app:sec:parallel_reasoning}), which is $[\mathrm{id}, \mathrm {parent}, \mathrm {conf.}, \mathrm {KV}^n, \mathrm {Seq}^{1 \sim n}]$, we need to concatenate the past KV caches and input sequences of different nodes into single large batch matrices. However, as discussed in Sec.~\ref{sec:motivation}, tree search paths exhibit varying path lengths, meaning that both past KV caches and context sequences have different sizes. 
To handle the length disparity and support arbitrary node parallelism, we apply padding for shorter past KV and context sequence: 
\begin{equation}
\label{eq:kv}
\begin{split}
    & \mathrm{KV}^{1\sim n} = \mathtt{concat}\left(\mathrm{KV}^{r^n}, \cdots, \mathrm{KV}^{a^n_1}, \mathrm{KV}^n\right), \\
    & \mathrm{padding}^n = \mathbf{0}_{\max\left(\forall_{m\in P}|\mathrm{KV}^{1\sim m}| \right) - |\mathrm{KV}^{1\sim n}|},  \\
    & \mathrm{KV}^{1\sim n}_{\mathrm{pad}} =  \mathtt{concat} \left(\mathrm{padding}^{n}, \mathrm{KV}^{1\sim n} \right), \\
    & \mathrm{KV}^{all} =  \mathtt{stack}\left ( [\forall_{n\in P}\ \mathrm{KV}^{1\sim n}_{\mathrm{pad}}]\right), 
\end{split}
\end{equation}
where $r$ and  $a_1, a_2, \dots$ are the root node and $1^{st}, 2^{nd}, \dots$ ancestors of $n$, respectively. $\mathbf{0}_{len}$ is a matrix of zeros with length $len$. Similar to above, the input sequences are also padded and stacked: 
\begin{equation}
\label{eq:seq}
\begin{split}
& \mathrm{padding}^n = padding\_token_{\max(\forall_{n\in P} |\mathrm{Seq}^{1\sim n}|)},  \\
& \mathrm{Seq^{1\sim n}_{pad}} =  \mathtt{concat} \left(  \mathrm{Seq^{1\sim n}}, \mathrm{padding}^n\right), \\ 
& \mathrm{Seq}^{all}=\mathtt{stack}\left( \forall_{n\in P}\mathrm{Seq^{1\sim n}_{pad}} \right), 
\end{split}
\end{equation}
where $padding\_token_{len}$ is a vector of the predefined padding token id with length $len$. 
In this way, the data is fed into the LLM for parallel generation. Additionally, techniques like DEFT~\cite{yao2024deft} are orthogonal to our DPTS and can be integrated to identify and merge shared prefixes, further optimizing inference efficiency.

\paragraph{Generation and Updating.} After the generation phase, we obtain new output sequences and past KV caches. These are then partitioned based on the tree width. Specifically, we segment past KV caches and only store those corresponding to new tokens. Output sequences are completely stored rather than fragmented, due to the negligible memory overhead. For clearer demonstration, the output of $i^\mathrm{th}$ node in $P$ can be written as 
\begin{center}
\begin{equation}
\label{eq:n_new}
\begin{split}
    & \mathbf{KV}^{n^{ij}} = \mathbf{n}.\mathtt{past\_kv}_{[ij, \dots, |\mathbf{KV}^{\text{all}}|:]}, \forall j \in \left[1, \dots, w\right] \\
    & \mathbf{Seq}^{1 \sim n^{ij}} = \mathbf{n}.\mathtt{output}_{[ij]} \\
    & n^{ij} = \left[\mathrm{id}, n^i, \mathbf{KV}^{n^{ij}}, \mathbf{Seq}^{1 \sim n^{ij}}\right] \\
    & \mathbf{n^{new}} = \left\{ n^{i1}, \dots, n^{iw} \right\}
\end{split}
\end{equation}
\end{center}
where $w$ is the tree width, $\mathbf{n}$ is the generation output in parallel manner. 
If sequences were stored in a fragmented manner, every inference step would require additional collection and concatenation, introducing unnecessary latency.  This approach is a trade-off between inference speed and memory consumption. 
Newly generated nodes are then updated into the candidate node pool $N$, making them available for subsequent selection processes.

\subsection{Algorithms for Searching and Transition Mechanism}
\label{sec:app:algorithms}

In the main text, due to paper length constraints, we only present the overall process in Algorithm~\ref{alg:algorithm_parallel}, which connects the entire framework’s algorithms and formulations, including Parallelism Streamline and the Search and Transition Mechanism.

In the above section, we provided the mathematical formulation of the Parallelism Streamline. In the following, we further supplement the algorithmic details of Search, Transition, and Reward, offering readers a clear and intuitive representation of the algorithmic process.

\begin{algorithm}[ht]
    \caption{Searching}
\label{alg:searching}
\begin{algorithmic}[1]
\REQUIRE Parallel queue $P$, current parallel queue size $\tau_P$, candidate node pool $N$. 
\ENSURE Updated $P$. 
\STATE $e_1\gets 0$
\FORALL{$n\in P$}
    \IF{$n$.mode$ = \mathtt{EXPLOIT}$}
    \STATE $e_1 \gets e_1  + 1$
    \ENDIF
\ENDFOR
\IF{$|P|<\tau_P$}
    \STATE $N'\gets$ Descending $N$ based on conf. \\
    \STATE $\mathbf{u} \gets N'[:\tau_P-|P|]$ 
    \FORALL{$u \in \mathbf{u}$}
        \IF {$ e_1<p|P| $}
        \STATE $u$.mode $\gets \mathtt{EXPLOIT}$
        \STATE $e_1 \gets e_1  + 1$
        \ELSE
        \STATE $u$.mode $\gets \mathtt{EXPLORE}$
        \ENDIF 
    \ENDFOR
    \STATE $P \gets P \cup \mathbf{u}$
\ENDIF
\RETURN $P$
\end{algorithmic}
\end{algorithm}

Firstly, Algorithm~\ref{alg:searching} demonstrates the searching mechanism. Specifically, during initialization or when the number of nodes in the parallel queue $P$ is less than the maximum parallelism $\tau_P$, the algorithm selects the $\tau_P- |P|$ highest-confidence nodes from the candidate node pool $N$ as supplementary nodes (Lines 8-9).

Then, the highest-confidence nodes are designated as exploit nodes (Line 12) until the proportion of exploit nodes reaches the ratio $p$. The remaining selected nodes are assigned as explore nodes. Finally, these newly selected nodes are merged into $P$ to prepare for the next cycle of parallel expansion. 

\begin{algorithm}[ht]
    \caption{Transition}
\label{alg:transition}
\begin{algorithmic}[1]
\REQUIRE Parallel queue $P$, transition thresholds $\theta_\mathrm{es}$ and $\theta_\mathrm{ds}$. 
\ENSURE Updated $P$. 
\FORALL{$n \in P$}
    \STATE $n^*=\max_\mathrm{conf.}$($n$.children)
    \IF{$n$.mode $=$ $\mathtt{EXPLOIT}$ \AND $c_{n^*} >\theta_\mathrm{es}$ \OR $n$.mode $=$ $\mathtt{EXPLORE}$ \AND $c_{n^*} >\theta_\mathrm{ds}$}
        \STATE $P \gets P\cup \{n^*\}$
        \STATE $n^*$.mode $\gets \mathtt{EXPLOIT}$ 
    \ENDIF
    \STATE $P \gets P \setminus n$
\ENDFOR
\RETURN $P$
\end{algorithmic}
\end{algorithm}

Next is the Transition Algorithm~\ref{alg:transition}. After each expansion, we iterate through all nodes in the parallel queue $P$ and identify the best child node  $n^*$ for each node.

Then, based on the category of node $n$, we compare $n^*$ with the corresponding threshold  $\theta$. If the early stop condition is not met or the deep seek condition is met, we add  $n^*$ as a new exploit node into $P$. Otherwise, the node will no longer be expanded and will be evicted from $P$.

\begin{algorithm}[ht]
    \caption{Reward}
\label{app:algo:reward}
\begin{algorithmic}[1]
\REQUIRE Parallel queue $P$, candidate node pool $N$. 
\ENSURE Updated $N$. 
\FORALL{$n \in P$} 
        \STATE $n$.children $\gets \text{Eq. (\ref{eq:n_new})}$ 
        \FORALL{$m \in n$.children}
            \IF{$\mathtt{is\_terminate}(m)$}
                \STATE $m$.reward $\gets \mathtt{reward}(m)$
            \ELSE
                \STATE $N \gets N \cup \{m\}$
            \ENDIF
        \ENDFOR
    \ENDFOR
\RETURN $N$
\end{algorithmic}
\end{algorithm}

After completing an expansion, we check whether each node’s path meets the termination conditions, such as reaching the maximum token limit or generating an end token. 
If a path satisfies the termination condition, exploration of that path stops, and a reward is computed as the final path score. We demonstrate this process in Algorithm~\ref{app:algo:reward}, which is identical to previous tree search algorithms and is not discussed in detail. However, for the sake of algorithmic completeness, we explicitly include it here.

\section{Additional Details about Experiment}
\label{app:sec:exp}

\subsection{Comparison Methods}
\label{app:sec:comparison_methods}

We compare DPTS against three widely used search algorithms: (1) Monte Carlo Tree Search~(MCTS)~\cite{sprueill2023monte} balances exploration and exploitation when sampling, while the selected paths rollout till termination, (2) Best-of-N~\cite{cobbe2021training} performs multiple independent rollouts and selects the highest-scoring output, (3) Beam Search~\cite{Yao_2023_Tree} expands multiple hypotheses in parallel, pruning low-scoring candidates at each step to maintain a fixed-width search~\cite{snell2024scaling}. 
Since efficient tree search algorithms have recently regained attention after the emergence of LLM reasoning, the strong baselines are limited. As a result, we primarily compare DPTS against these typical and well-established search algorithms to demonstrate the effectiveness of our method.

\subsection{Experimental Settings}
\label{app:sec:exp_setting}


The experimental settings are as follows: we set the tree width to 4, tree depth to 16, mini step to 100, and the maximum token limit to 2048. The MCTS time limit is 120 seconds, and the threshold parameter is empirically set to $t^*=5$. All models were downloaded from Hugging Face.

For evaluation, we implemented a custom codebase, which is included in the supplementary materials. Inference automatically terminates if it exceeds the timeout limit or encounters a memory overflow. Within these constraints, the search tree can expand and roll out indefinitely, ensuring comprehensive exploration during inference. 

\subsection{Distribution of Best Path Index}
\label{app:sec:best_path_index}

We use a histogram to visualize the earliest (blue bar) and shortest (green bar) best path index. Through an ablation study, we examine how the best solution in the search path evolves as our method is progressively introduced.

In the baseline method, the best path typically appears around the 8th terminated path. Incorporating the parallelism streamline does not directly affect the accuracy of the search path. However, after adding the searching and transition mechanism, DPTS finds the best solution region more quickly and reaches the best path earlier.

\begin{figure}
    \centering
    \includegraphics[width=\linewidth]{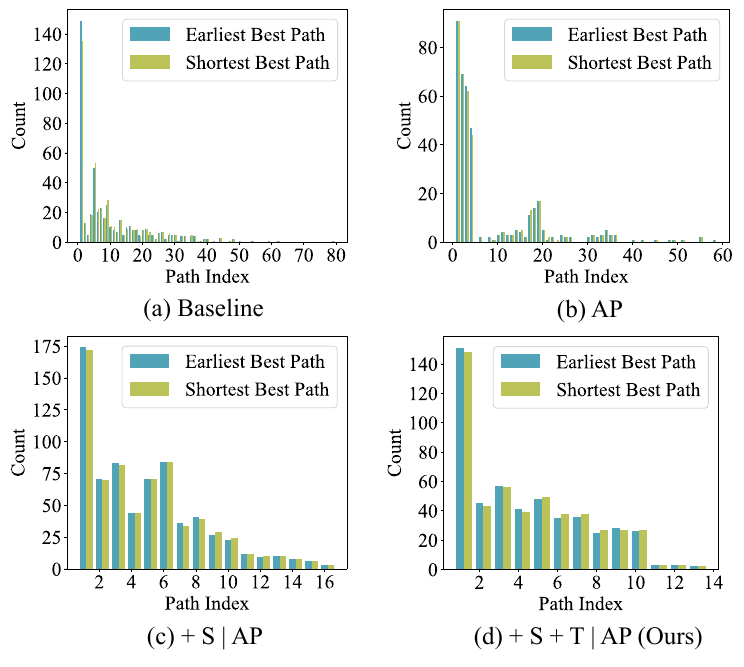}
    \caption{The distribution of the earliest (blue bar) and shortest (green bar) best path index. }
    \label{fig:best_path_index}
\end{figure}

\subsection{Additional Results of $\lambda$}
\label{sec:app:hyperparameter}

\begin{figure}
    \centering
    \includegraphics[width=\linewidth]{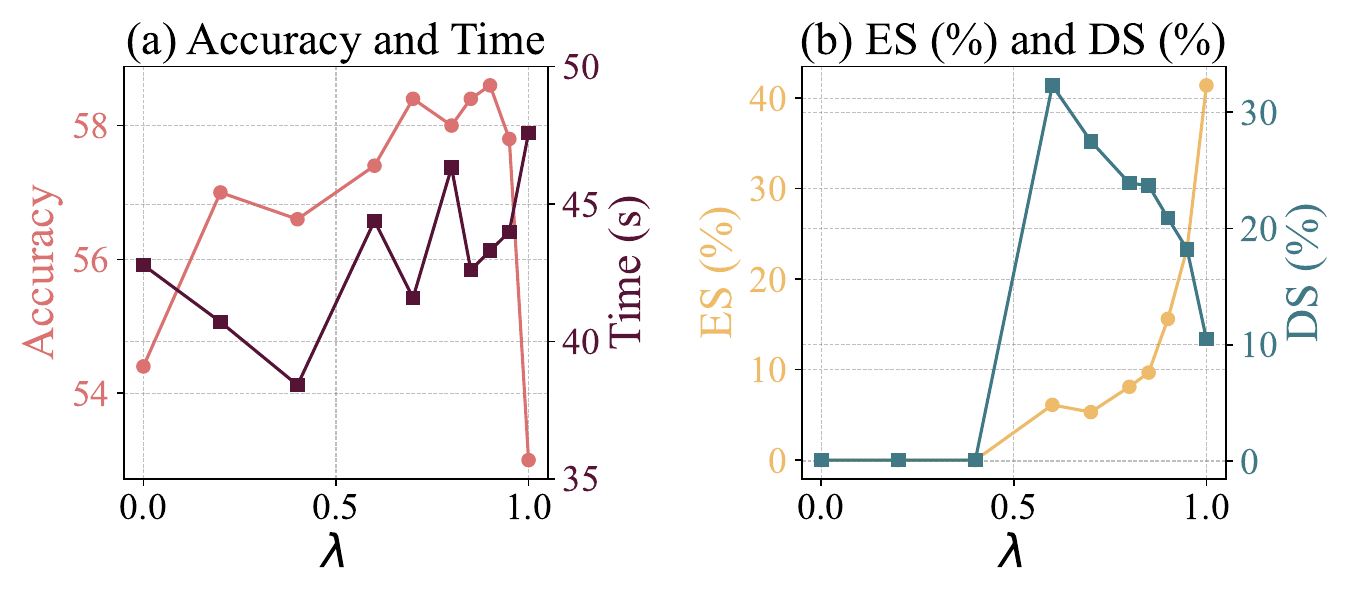}
    \caption{Illustrations of hyperparameter analysis. }
    \label{app:fig:lambda_curve}
\end{figure}

We conducted a more detailed experiment on $\lambda$, with results presented in Table~\ref{app:tab:hyperparameter} and Figure~\ref{app:fig:lambda_curve}. It is evident that when $\lambda$ is set within a reasonable range (e.g., $[0.7,0.9]$), both accuracy and inference time exhibit optimal performance. In this range, the proportion of DS\% (deep seek transitions) is higher than ES\% (early stop transitions), indicating that more high-confidence nodes are being timely converted into exploit nodes. At the same time, a small number of paths are reassigned as low-score paths during inference and subsequently terminated.

However, when $\lambda$ is too large (e.g., close to $1$), the proportion of es increases aggressively. This suggests that many exploit nodes being expanded have scores within the range of $[0.9,1.0]$, and setting the threshold in this range may cause some correct paths to be prematurely stopped. Conversely, when $\lambda$ is too small (e.g., $< 0.4$), both ES and DS proportions drop to nearly zero. This occurs because most node scores exceed the threshold, causing nearly all paths to expand under exploitation mode. Moreover, an overly small early stop threshold causes no paths to terminate, effectively degrading the search into an exploitation-only strategy. 

Therefore, selecting a suitable $\lambda$ is important. A larger  $\lambda$ imposes stricter exploitation conditions, leading to more paths being stopped and fewer paths being converted to deep seeking. Conversely, a smaller $\lambda$ results in looser conditions, allowing more exploiting paths to continue rolling out until they reach a termination condition, while more high-confidence paths transition into deep seeking.

\begin{table}
    \centering
    \caption{Hyperparameter $\lambda_\mathrm{es}$ and $\lambda_\mathrm{ds}$ in transition thresholds $\theta_\mathrm{es}$ and $\theta_\mathrm{ds}$. ``ES (Early Stop)~\%'' and ``DS (Deep Seek)~\%'' are the ratios of the node type transition between the exploitation and exploration.  }
    \label{app:tab:hyperparameter}
    \resizebox{\linewidth}{!}{
    \begin{tabular}{cccccc}
    \toprule
        \textbf{$\lambda_\mathrm{es}$} & \textbf{$\lambda_\mathrm{ds}$} & \textbf{Acc.} & \textbf{Time (s)} & \textbf{ES~(\%)}  & \textbf{DS~(\%)} \\ \midrule
        1.0 & 1.0 & 53.0 & 47.59 & 41.4 & 10.5 \\ 
        0.95 & 0.95 & 57.8 & 43.99 & 23.4 & 18.2 \\ 
        0.9 & 0.9  & 58.6 & 43.30 & 15.6 & 20.9 \\
        0.85 & 0.85 & 58.4 & 42.60 & 9.67 & 23.7 \\
        0.8 & 0.8 & 58.0 & 46.33 & 8.1  & 23.9 \\
        0.7 & 0.7 & 58.4 & 41.58 & 5.3 & 27.5 \\
        0.6 & 0.6 & 57.4 & 44.39 & 6.1 & 32.3   \\
        0.4 & 0.4 & 56.6 & 38.41 & 0 & 0.1 \\
        0.2 & 0.2 & 57.0 & 40.71 & 0 & 0.1  \\
        0   & 0   & 54.4 & 42.78 & 0 & 0.1  \\
        \bottomrule
    \end{tabular}
    }
    \vspace{5pt}
\end{table}

\section{Related Work}
\label{app:sec:related_work}

\paragraph{Reasoning with LLMs.}
The success of ChatGPT~\cite{achiam2023gpt} has driven significant interest in Transformer-based LLMs~\cite{bai2023qwen,touvron2023llama,yang2024qwen2}, initially applied to simple System 1 tasks like translation and summarization~\cite{Brown_2020_Language,Ouyang_2022_Training}. As LLM capabilities grew, research shifted toward enhancing their ability to handle more complex System 2 tasks, such as mathematical reasoning and logic~\cite{Kojima_2022_Large,hao2023reasoning,zhang2024accessing,hosseini2024v}. The introduction of Chain-of-Thought~(CoT)~\cite{Wei_2022_Chain} advanced this domain by breaking down problems into intermediate steps, which proved effective for multi-step reasoning~\cite{Kojima_2022_Large}. Recent research~\cite{zhang2022automatic, bi2024forest} has proposed numerous variations, such as Self-Consistent CoT~\cite{wang2022self}, R-CoT~\cite{deng2024r} to improve its ability,  but its exploration scope remains constrained, limiting its effectiveness.~~\cite{chu2023survey}.

\paragraph{Tree Search for Reasoning.}
To address the limitations of linear reasoning in CoT, the Tree of Thoughts~(ToT)~\cite{Yao_2023_Tree} framework was introduced, which aligns with the inference scaling law, showing that increasing Test-Time Compute can enhance LLM reasoning abilities~\cite{snell2024scaling,wu2024inference}.
The simplest form of tree search, Best-of-N, samples multiple reasoning paths and selects the best one~\cite{ cobbe2021training,lightman2023let,jiao2024learning}. Beam search extends this approach by considering multiple paths in parallel~\cite{Yao_2023_Tree}. Recent work~\cite{hao2023reasoning} has leveraged the exploration capabilities of Monte Carlo Tree Search~(MCTS)~\cite{chaslot2008monte, kocsis2006improved, browne2012survey}. For instance, MCTS-rollout~\cite{wan2024alphazero} introduces backtracking to explore different paths, while Q${^*}$~\cite{wang2024q} uses heuristic functions to guide rollout, and ReST-MCTS${^*}$~\cite{zhang2024rest} uses MCTS to sample traces for self-training. 
However, challenges remain in its computational efficiency, with limited research on accelerating tree search for LLM reasoning.

\paragraph{LLM Inference Acceleration.}
LLMs face efficiency bottlenecks during inference, leading to significant efforts aimed at accelerating inference~\cite{lin2024awq,hong2024flashdecoding++,leviathan2023fast,fu2024break,jing2023deep,cai2024medusa}. However, most of these approaches are designed for linear decoding tasks and are not directly applicable to tree-structured reasoning~\cite{li2024large, zhou2024survey}. 
Recent work~\cite{ning2023skeleton, han2024token,nayab2024concise}, such as Deft ~\cite{yao2024deft} focuses on optimizing common prefixes in tree reasoning using operator-level enhancements, , while others leverage self-consistency to enable early stopping~\cite{li2024escape, wan2024dynamic}. Despite such advancements, building efficient tree search algorithms for reasoning remains an under-explored area. The need for specialized algorithms that can scale with complex tree-structured reasoning, while maintaining efficiency, remains a key open challenge in LLM inference acceleration.






\end{document}